\documentclass[10pt,twocolumn,letterpaper]{article}

\usepackage[pagenumbers]{cvpr} 

\usepackage[table]{xcolor}
\usepackage{booktabs}

\definecolor{greenLight}{HTML}{E8F8F2}  
\definecolor{greenStrong}{HTML}{A9DFBF} 

\definecolor{redLight}{HTML}{FDEDEC}    
\definecolor{redStrong}{HTML}{F5B7B1} 

\definecolor{cvprblue}{rgb}{0.21,0.49,0.74}
\usepackage[pagebackref,breaklinks,colorlinks,allcolors=cvprblue]{hyperref}

\def\MethodName{DynaVid}

\usepackage{kotex}
\usepackage[capitalize]{cleveref}
\usepackage{multirow}
\Crefname{section}{Sec.}{Secs.}
\Crefname{section}{Section}{Sections}
\Crefname{table}{Table}{Tables}
\crefname{table}{Tab.}{Tabs.}
\usepackage[table]{xcolor}

\def\paperID{2269} 
\def\confName{CVPR}
\def\confYear{2026}

\title{Learning to Generate Highly Dynamic Videos using Synthetic Motion Data}

\author{
Wonjoon Jin$^{1}$ \quad
Jiyun Won$^{1}$ \quad
Janghyeok Han$^{1}$ \quad
Qi Dai$^{2}$ \\
Chong Luo$^{2}$ \quad
Seung-Hwan Baek$^{1}$ \quad
Sunghyun Cho$^{1}$ \\[0.6em]
\large $^{1}$POSTECH \qquad $^{2}$Microsoft Research Asia
}

\begin{document}

\twocolumn[{%
\renewcommand\twocolumn[1][]{#1}%
\maketitle
\vspace{-2.0em}
\centering
\includegraphics[width=\linewidth]{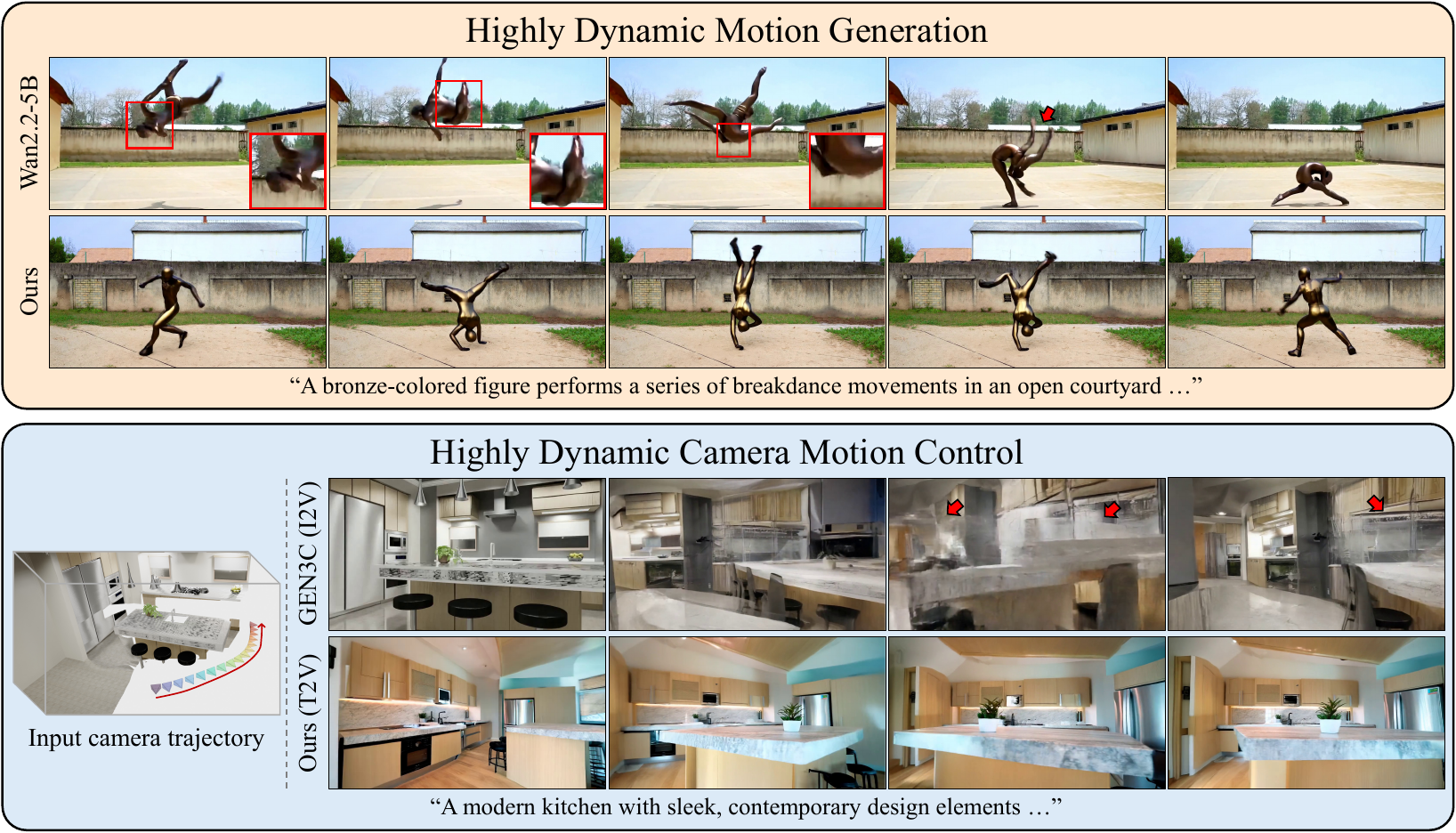}
\vspace{-7.0mm}
\captionof{figure}{
Examples of video synthesis results for highly dynamic object motion (top) and camera-controlled video generation with rapid viewpoint changes (bottom).
Our method produces natural and highly dynamic motions, whereas Wan2.2-5B~\cite{wan2025wan} generates unrealistic motion and GEN3C~\cite{ren2025gen3c} exhibits noticeable visual artifacts. 
The synthesis results are best viewed in the supplementary video.
}
\vspace{4.0mm}
\label{fig:teaser}
}]

\begin{abstract}

\vspace{-3.0mm}

Despite recent progress, video diffusion models still struggle to synthesize realistic videos involving highly dynamic motions or requiring fine-grained motion controllability. 
A central limitation lies in the scarcity of such examples in commonly used training datasets. 
To address this, we introduce \MethodName{}, a video synthesis framework that leverages synthetic motion data in training, which is represented as optical flow and rendered using computer graphics pipelines. 
This approach offers two key advantages. 
First, synthetic motion offers diverse motion patterns and precise control signals that are difficult to obtain from real data. 
Second, unlike rendered videos with artificial appearances, rendered optical flow encodes only motion and is decoupled from appearance, thereby preventing models from reproducing the unnatural look of synthetic videos.
Building on this idea, \MethodName{} adopts a two-stage generation framework: a motion generator first synthesizes motion, and then a motion-guided video generator produces video frames conditioned on that motion.
This decoupled formulation enables the model to learn dynamic motion patterns from synthetic data while preserving visual realism from real-world videos. 
We validate our framework on two challenging scenarios, vigorous human motion generation and extreme camera motion control, where existing datasets are particularly limited.
Extensive experiments demonstrate that \MethodName{} improves the realism and controllability in dynamic motion generation and camera motion control.

\end{abstract}

\section{Introduction}
\label{sec:intro}

Highly dynamic motions, such as breakdancing and rapid camera movements, are prominent elements that enhance the visual impact in modern video content, including films, animations, and social media. 
However, despite recent progress in video generation~\cite{blattmann2023stable, polyak2024movie, yang2024cogvideox, wan2025wan}, synthesizing realistic videos with such motions remains challenging (\cref{fig:teaser}).
A primary bottleneck lies in the training data: although existing video diffusion models are trained on extremely large-scale datasets~\cite{pexels_dataset, bain2021frozen, chen2024panda}, video clips containing highly dynamic motions are relatively underrepresented.
Moreover, manually collecting such videos is labor-intensive and difficult to scale, making it challenging to construct balanced datasets that adequately capture dynamic motion.

Another challenge is controllability, i.e., the ability to guide a model to generate videos with specific target motions.
For example, controlling rapidly changing viewpoints during video synthesis remains particularly difficult.
Existing camera-controllable video diffusion models~\cite{wang2024motionctrl,he2025cameractrl,bahmani2025ac3d} typically require accurate ground-truth 3D camera poses during training.
However, estimating accurate camera poses for videos with extreme camera motion is highly unreliable due to minimal frame overlap.
As a result, these models are usually trained on relatively simple datasets with limited camera motions, leading to degraded performance when synthesizing complex camera motions (\cref{fig:teaser}).

A straightforward way to address these dataset limitations is to use synthetic data.
Synthetic data generated from computer graphics pipelines provides dynamic motion scenes and precise control signals, such as camera parameters, which are difficult to capture in practice.
Recent studies have explored training video diffusion models on rendered synthetic videos to improve motion fidelity and controllability~\cite{zhao2025synthetic, shuai2025free}.
However, building synthetic video datasets with both highly dynamic motions and natural appearances is itself a formidable task.
Rendered videos often contain artificial textures, lighting, and shadows, creating a large domain gap from real footage.
As a result, models trained on such datasets tend to reproduce the artificial look of synthetic videos rather than realistic visuals.

To more effectively harness the advantages of synthetic data while mitigating its drawbacks, we propose \MethodName{}, a video synthesis framework that leverages \emph{synthetic motion data}, represented as optical flow and rendered from computer graphics engines, instead of synthetic videos.
Unlike videos, optical flow encodes only motion information and is naturally decoupled from appearance, thereby substantially reducing the domain gap from real motion.
We render synthetic motion data along with precise control signals and use them to train video diffusion models.
This approach enables the model to learn highly dynamic motion patterns and fine-grained controllability without sacrificing visual realism.

Based on this idea, \MethodName{} is designed as a two-stage generation framework: a motion generator synthesizes optical flow maps and a motion-guided video generator produces RGB video frames from the generated flow.
We train this framework using both synthetic and real-world datasets.
The synthetic motion data provides supervision for the motion generator, allowing it to learn diverse and highly dynamic motion patterns.
In parallel, optical flows are estimated from real-world videos to construct motion-video pairs, which are then used to train the motion-guided video generator.
Overall, our decoupled framework, equipped with the complementary strengths of synthetic motion data and real-world videos, results in natural-looking video synthesis that faithfully captures highly dynamic motions.

Our framework can be further extended to incorporate controllability.
We integrate a control branch into the motion generator to condition the generation process on additional control signals.
For example, by training on paired data of synthetic motion and corresponding camera parameters, the motion generator learns to produce flow maps reflecting rapidly changing camera motions.
These generated motion then provides control signals to the motion-guided video generator, enabling camera-controlled video synthesis with complex viewpoint change.

We demonstrate the effectiveness of our framework with two particularly challenging scenarios: (1) vigorous human motions such as breakdancing, and (2) camera-controlled video synthesis with rapid camera motions.
Extensive experiments show that our approach can generate natural-looking videos with highly dynamic motions and precise motion controllability for extreme scenarios.
Our main contributions are summarized as follows:
\begin{itemize}
    \item We construct synthetic datasets that capture dynamic motion scenes and provide precise control signals that are difficult to obtain from real-world videos.
    \item Instead of rendered videos with artificial appearances, we leverage rendered optical flow for training, which aligns closely with real motion and avoids appearance gaps. This enables the model to learn dynamic motion patterns from synthetic data while preserving visual realism from real videos through a decoupled, two-stage framework.
    \item Extensive experiments demonstrate that our method outperforms existing approaches in terms of dynamic motion generation and fine-grained camera motion control.
\end{itemize}

\section{Related Work}
\label{sec:related}

\vspace{-0.5mm}
\paragraph{Video diffusion models.}
Recent video diffusion models~\cite{peebles2023scalable, blattmann2023stable, polyak2024movie, yang2024cogvideox, wan2025wan, videoworldsimulators2024, sora2, kling} have achieved remarkable progress in text-to-video synthesis.
Leveraging large-scale datasets and advanced diffusion frameworks~\cite{ho2020denoising, lipman2022flow}, they successfully learn spatio-temporal structures of real videos.
However, open-source models such as CogVideoX and Wan~\cite{yang2024cogvideox, wan2025wan} still struggle to generate highly dynamic scenes, primarily due to the limited diversity of motion in available training data.

\vspace{-4mm}
\paragraph{Object motion synthesis.}
Synthesizing and controlling object motion is crucial in content creation and visual storytelling.
Recent video diffusion models leverage large-scale video corpora paired with detailed text descriptions, enabling realistic motion generation~\cite{wan2025wan, videoworldsimulators2024, sora2}.
Another line of work introduces explicit motion control using additional conditioning signals such as 2D pose sequences or motion trajectories~\cite{jain2024peekaboo, yang2024direct, geng2025motion, burgert2025go, namekata2024sg, shi2024motion, wang2024motionctrl, wu2024draganything, chen2023motion, yin2023dragnuwa, hu2024animate, xu2024magicanimate, zhang2024mimicmotion, xu2025hypermotion, zhang2023adding}.
For instance, Animate Anyone~\cite{hu2024animate} guides human motion generation using pose sequences, while Go-with-the-Flow~\cite{burgert2025go} warps latent noise volumes according to optical flow.
However, these methods often struggle to synthesize highly dynamic motions due to the scarcity of such examples in their training datasets.

\vspace{-4mm}
\paragraph{Camera-controlled video synthesis.}
Controlling camera motion during video synthesis has recently gained significant attention~\cite{he2025cameractrl, wang2024motionctrl, jin2025flovd, bahmani2025ac3d, bahmani2024vd3d, genie3, ren2025gen3c, gu2025diffusion}.
These models are typically trained on paired datasets of video frames and corresponding camera parameters so that the generated videos can follow specified camera trajectories.
CameraCtrl~\cite{he2025cameractrl} represents camera rays using Plücker embeddings computed from camera parameters, while VD3D~\cite{bahmani2024vd3d} and AC3D~\cite{bahmani2025ac3d} extend this idea with Diffusion Transformers.
FloVD~\cite{jin2025flovd} and GEN3C~\cite{ren2025gen3c} leverage depth estimation to enable 3D-aware video generation with accurate camera control.
However, since these models are trained mostly on videos with limited viewpoint variation, they still struggle to synthesize videos under extreme camera motions, such as rapid rotations or large translations.

\vspace{-4mm}
\paragraph{Training with synthetic data.}
Synthetic data created from rendering engines~\cite{blender2018, unrealengine} have long been employed for dense prediction tasks~\cite{dosovitskiy2015flownet, teed2020raft, ranftl2020towards, mayer2016large, greff2022kubric, butler2012naturalistic, wang2025waft} and is increasingly explored in generative modeling~\cite{sharma2024alchemist, zhao2025synthetic, liu2025revision, shuai2025free, bai2025recammaster}.
In generative settings, Sharma et al.~\cite{sharma2024alchemist} train models on paired rendered images and physical parameters for controllable material synthesis.
Zhao et al.~\cite{zhao2025synthetic} leverage rendered videos to learn natural object motion generation, while Shuai et al.~\cite{shuai2025free} employ rendered videos with full 6D annotations of object and camera poses for joint object-camera control.
However, models trained directly on rendered videos often suffer from unrealistic appearances due to the substantial visual discrepancy between rendered and real footage.
In contrast, we leverage synthetic motion data represented as optical flow instead of rendered videos, allowing the model to learn complex motion dynamics while effectively mitigating the appearance domain gap between synthetic and real videos.

\vspace{-4mm}
\paragraph{Multi-Modal generation.}
Diffusion models have shown strong capability in modeling cross-modal relationships across diverse domains.
In video generation, a few prior studies exploit dual modalities (e.g., optical flow and RGB video) through two-stage pipelines~\cite{liang2024movideo, ni2023conditional, shi2024motion, jin2025flovd, pandey2025motion}.
For instance, MoVideo~\cite{liang2024movideo} and LFDM~\cite{ni2023conditional} first synthesize optical flow maps and subsequently use them to guide conditional video generation.
However, these methods primarily rely on real-world data and offer limited motion controllability.
In contrast, our approach introduces a control branch to enhance motion controllability and employs synthetic motion data to learn highly dynamic motion patterns beyond what real datasets provide.

\begin{figure}[!t]
    \centering
    \includegraphics[width=\linewidth]{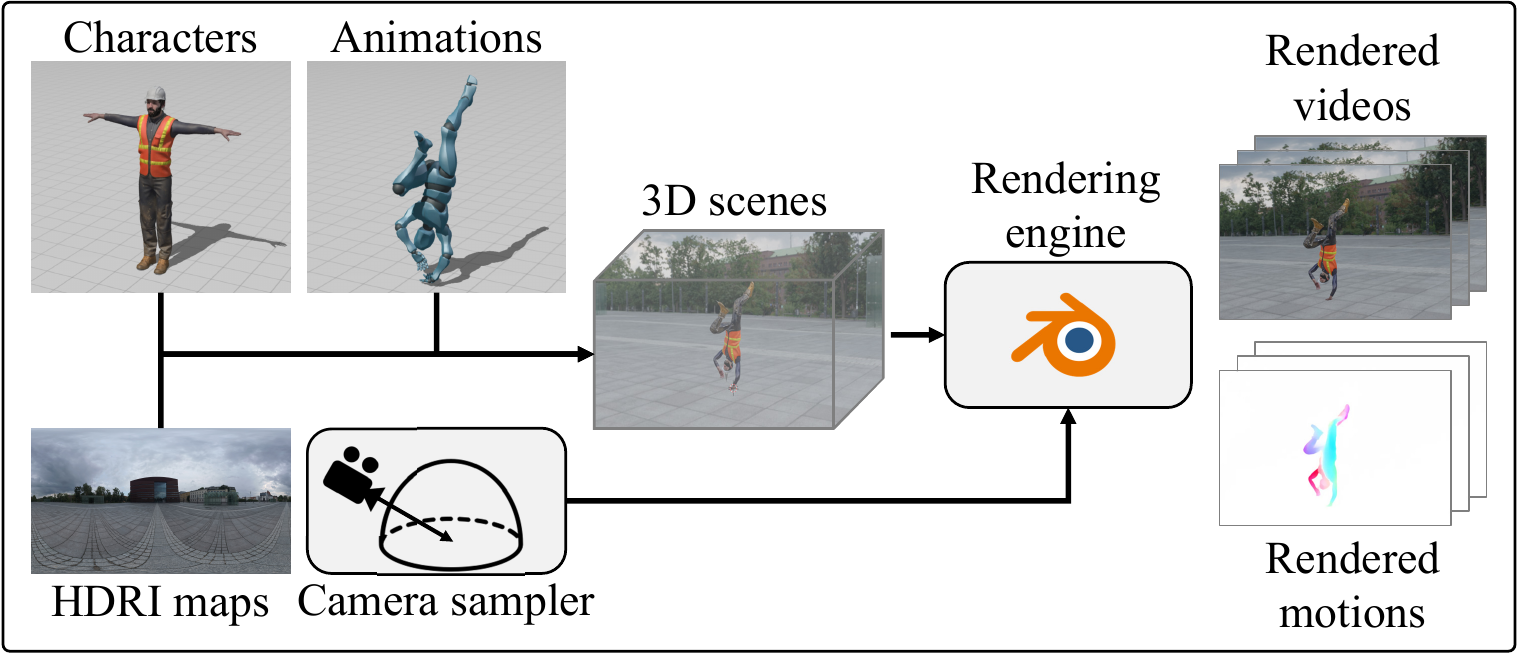}
    \vspace{-6.5mm}
    \caption{
    Overview of our dataset generation pipeline.
    }
    \label{fig:data}
    \vspace{-3mm}
\end{figure}

\begin{figure*}[!t]
    \centering
    \includegraphics[width=0.98\linewidth]{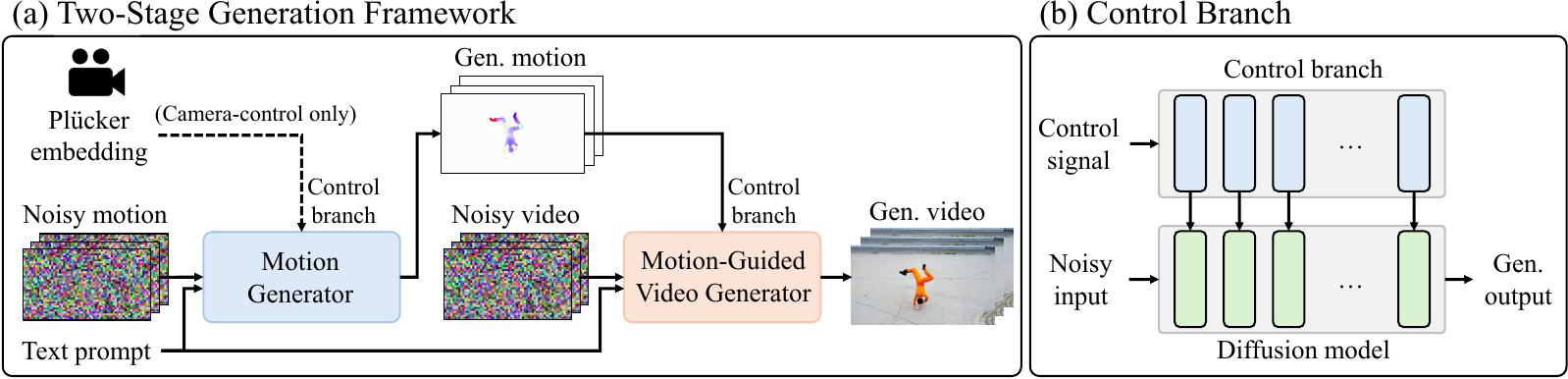}
    \vspace{-3mm}
    \caption{
    Overview of \MethodName{}. (a) The motion generator first synthesizes motion and then produces video frames conditioned on the generated motion. For camera-controlled video synthesis, Pl\"ucker embeddings are provided as additional input. (b) Our framework adopts VACE~\cite{jiang2025vace} to incorporate control signals such as Pl\"ucker embeddings or optical flow maps.
    }
    \vspace{-4mm}
    \label{fig:overview}
\end{figure*}

\section{Method}
We propose a novel approach that leverages synthetic motion data to enable highly dynamic video generation.
Our approach particularly targets two challenging scenarios: vigorous human motions like breakdancing and dynamic camera control involving rapid viewpoint changes.
To achieve this, we first construct a dataset generation pipeline that produces synthetic motion data using computer graphics pipelines.
We then introduce DynaVid, our two-stage video generation framework equipped with a training strategy using both synthetic and real data.

\subsection{DynaVid Datasets}
\label{sec:syn_data_gen}
We construct a dataset generation pipeline based on the Cycles renderer in Blender~\cite{blender2018} and use it to synthesize two datasets: \textit{DynaVid-Human} and \textit{DynaVid-Camera}.
The pipeline consists of three stages: 3D scene construction, camera trajectory definition, and rendering (\cref{fig:data}).

\vspace{-3mm}
\paragraph{DynaVid-Human.}
We first construct 3D scenes using publicly available assets, where vigorous human actions are represented by integrating animatable human characters with motion sequences from the Mixamo dataset~\cite{mixamo}.
We then define camera trajectories $\mathcal{C}^{syn}=\{{C^{syn}_{n}}\}_{n=1}^{N}$, where $n$ denotes the frame index.
For DynaVid-Human, a single camera position is used across all frames, randomly sampled on a hemisphere and fixed while oriented toward its center.
Finally, we render synthetic RGB videos and their corresponding optical flow maps, $\mathcal{I}^{syn}=\{{I^{syn}_{n}}\}_{n=1}^{N}$ and $\mathcal{F}^{syn}=\{{f^{syn}_{n}}\}_{n=1}^{N}$, using Blender’s Cycles renderer.
Optical flow is computed by measuring the 3D displacement of each visible surface point between consecutive frames and projecting it onto the image plane, yielding pixel-wise motion vectors.
These rendered flows serve as synthetic motion data used to train the motion generator in our framework.
For simplicity, we omit the frame index $n$ in the following.

\vspace{-3mm}
\paragraph{DynaVid-Camera.}
We adapt the same pipeline with minor modifications to simulate dynamic camera motion scenarios.
Specifically, we construct complex 3D environments (e.g., urban and natural scenes) using diverse mesh assets from BlendSwap~\cite{blendswap}.
Camera trajectories are defined by preselecting several key camera positions in each scene and interpolating them using Non-Uniform Rational B-Spline (NURBS) curves to produce highly dynamic trajectories with rapid viewpoint changes.
Additional details on both datasets are provided in the supplementary document.

\subsection{Video and Motion Representations}

Following latent diffusion models~\cite{rombach2022high}, we represent both videos and motions in the latent space of a pre-trained Variational Auto-Encoder (VAE). To reuse the VAE trained on RGB videos~\cite{wan2025wan} for optical flow, we convert the flow sequences into the RGB domain via HSV mapping~\cite{chefer2025videojam}.

Specifically, each pixel’s flow vector $f(\text{p})$ is first normalized so that the flow magnitudes are more evenly distributed within the $[0, 1]$ range while preserving direction:
\begin{equation}
\begin{aligned}
f'(\text{p}) = s(f(\text{p}))\frac{f(\text{p})}{\lVert f(\text{p}) \rVert_{2}},~
s(f(\text{p})) = \min\left(1,\sqrt{\frac{\lVert f(\text{p}) \rVert_{2}}{s_f}}\right),
\end{aligned}
\label{eq:flow_normalization}
\end{equation}
where $\lVert \cdot \rVert_{2}$ denotes the L2 norm and $s_f$ is a dataset-level scale factor defined as the 99th percentile of flow magnitudes.
The magnitude $m = \lVert f'(\text{p}) \rVert_{2}$ and the angle $\alpha = \text{arctan2}(f'(\text{p})_{y}, f'(\text{p})_{x})$ of the normalized flow are mapped to the value and hue channels in HSV space, respectively, with saturation fixed to 1.
The resulting HSV image is then converted into RGB format and encoded to the latent space by the VAE.
Further details and discussion of this conversion process are provided in the supplementary document.
For simplicity, we omit the notation of the latent representations for video and motion in the following sections.

\subsection{DynaVid Framework}
\label{sec:framework}

\cref{fig:overview} (a) illustrates the overall two-stage video generation framework.
This framework decouples motion synthesis from video synthesis: the motion generator synthesizes motion represented as optical flow, and the motion-guided video generator produces RGB videos conditioned on this motion.
This decoupled design enables the use of synthetic motion data while avoiding the visual domain gaps that arise from synthetic videos.
We build both stages on top of a recent video diffusion model~\cite{wan2025wan}, so the framework is largely compatible with existing diffusion-based generators.

\vspace{-3.1mm}
\paragraph{Motion generator.}
Given a noisy motion $\mathcal{F}_{t}$ and a text condition $c_{\text{txt}}$, the motion generator iteratively denoises $\mathcal{F}_{t}$ to obtain a clean motion $\mathcal{F}$, represented as optical flow maps. This stage is responsible only for deciding “how things move,” not for rendering appearance.
To enable camera-controlled motion generation, we extend the motion generator with a control branch following a conditional video diffusion architecture, e.g., VACE~\cite{jiang2025vace}, as shown in~\cref{fig:overview} (b). Specifically, the control branch receives Pl\"ucker-embeddings computed from camera parameters $\mathcal{C}$ as input and produces context feature maps. These features are injected into the transformer blocks of the motion generator through pixel-wise addition, so that the denoising process can produce flow maps that are consistent with the given camera motion. In this way, camera trajectories become an explicit control signal for the motion stage.

\vspace{-3.1mm}
\paragraph{Motion-guided video generator.}
The motion-guided video generator takes a noisy video $\mathcal{I}_{t}$ along with the text condition $c_{\text{txt}}$ and generated motion $\mathcal{F}$, and denoises it to produce the final video $\mathcal{I}$. We again adopt the conditional video diffusion architecture~\cite{jiang2025vace} to inject control signals. The overall structure is similar to that of the motion generator, except that the control signal is the optical flow, rather than the camera parameters.
Because optical flow captures both object and camera motions, and the motion-guided video generator is explicitly trained to follow the provided flow, the same architecture applies to both dynamic object motion and camera-controlled video synthesis.

\subsection{Training \MethodName{}}
\label{sec:training}

We train \MethodName{} using both synthetic and real-world datasets.
During training, real-world data provide natural appearance and general motion statistics, whereas synthetic data supply precisely controlled and highly dynamic motion patterns.
This combination enables the model to learn realistic video generation while capturing a wide spectrum of motion dynamics.

\vspace{-3mm}
\paragraph{Motion generator.}
The motion generator is trained from two motion sources: synthetic motion data $\mathcal{F}^{\text{syn}}$ and real motion data $\mathcal{F}^{\text{real}}$.
For $\mathcal{F}^{\text{real}}$, we extract optical flow from real videos using an off-the-shelf flow estimator~\cite{wang2025waft}. We use an internal video dataset, containing scenes similar to those in the Pexels dataset~\cite{pexels_dataset}, for dynamic-object scenarios, and the RealEstate10K dataset (RE10K)~\cite{zhou2018stereo} for camera-controlled scenarios.
For $\mathcal{F}^{\text{syn}}$, we use rendered motion data from DynaVid-Human and DynaVid-Camera, where ground-truth optical flow is directly obtained from the renderer (\cref{sec:syn_data_gen}).

Training proceeds in two stages.
We first pretrain the motion generator on $\mathcal{F}^{\text{real}}$ to learn general, in-the-wild motion statistics.
We then fine-tune it with $\mathcal{F}^{\text{syn}}$ to expand its capability toward highly dynamic motions.
During fine-tuning, each batch contains a mixture of real and synthetic flows so that the model does not forget natural motions while acquiring extreme ones.
We use the Flow Matching objective~\cite{lipman2022flow} in both training stages:
\begin{equation}
\begin{aligned}
\mathbb{E}_{\mathcal{F},c_{\text{txt}},{C},\epsilon,t_f} &\big[\lVert \hat{u}^{\mathcal{F}}(\mathcal{F}_{t_{f}}; c_{\text{txt}}, {C}, t_f) - v^{\mathcal{F}}\rVert^{2}_{2} \big],
\label{eq:loss_motion_generator}
\end{aligned}
\end{equation}
where $\mathcal{F}_{t_{f}}=(1-t_f)\mathcal{F}_0+t_f\mathcal{F}_1$, $v^{\mathcal{F}} = \epsilon - \mathcal{F}$ is the target velocity and $\epsilon \sim \mathcal{N}(0, I)$.
The camera parameter $C$ is used only for the camera-controlled setting.

\vspace{-3mm}
\paragraph{Motion-guided video generator.}
The motion-guided video generator is trained on real-world paired datasets of video frames $\mathcal{I}^{real}$ and their corresponding optical flow maps $\mathcal{F}^{real}$, where $\mathcal{F}^{real}$ are extracted from $\mathcal{I}^{real}$ as described above.
We use our internal dataset for both dynamic-object and camera-controlled scenarios.
This supervision guides the model to synthesize realistic visual appearances while faithfully following the input motion.

To further improve motion fidelity, we introduce a simple yet effective dataset filtering technique.
Since real-world motion-video pairs are constructed using estimated optical flow, they inevitably contain estimation errors.
We quantify these errors via flow cycle consistency computed from forward and backward flows, where the maximum observed error reaches 1080.05 pixels.
To mitigate this issue, we discard motion-video pairs with large consistency errors by applying a threshold of 1.19 pixels, corresponding to the 90th percentile of the consistency error distribution.
This filtering reduces the influence of inaccurate flow supervision and ensures that the model faithfully follows the input motion.

The motion-guided video generator is trained with the Flow Matching objective~\cite{lipman2022flow}:
\begin{equation}
\begin{aligned}
\mathbb{E}_{\mathcal{I},\mathcal{F},c_{\text{txt}},\epsilon, t_I} &\big[\lVert \hat{u}^{\mathcal{I}}(\mathcal{I}_{t_{I}}; c_{\text{txt}}, \mathcal{F}, t_I) - v^{\mathcal{I}}\rVert^{2}_{2} \big],
\label{eq:loss_video_generator}
\end{aligned}
\end{equation}
where $\mathcal{I}_{t_{I}}=(1-t_I)\mathcal{I}_0+t_I\mathcal{I}_1$ and $v^{\mathcal{I}}=\epsilon-\mathcal{I}$ denotes the target velocity for the video representation.
Note that camera parameters are not used in this stage, as camera control is already encoded in the motion generated by the motion generator.
This training scheme enables the model to synthesize natural-looking videos while capturing highly dynamic motions that are rare or absent in real-world data.

\begin{figure*}[!t]
    \centering
    \includegraphics[width=\linewidth]{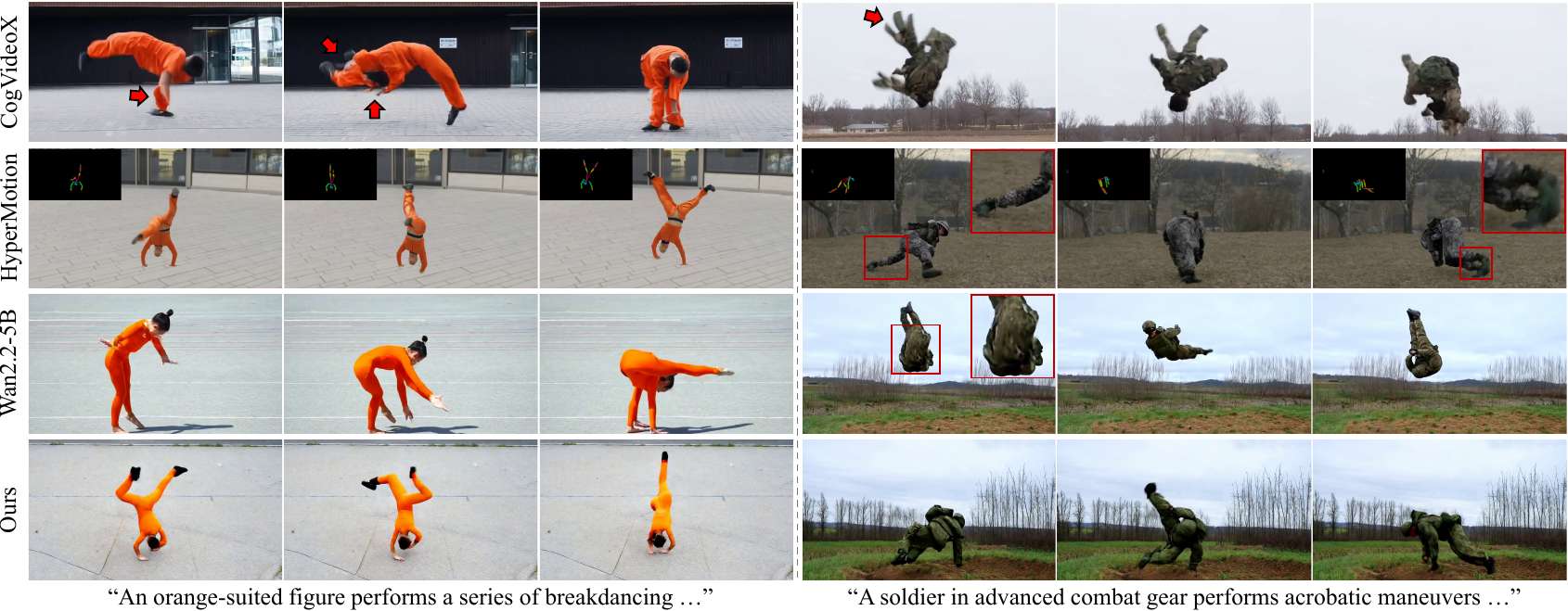}
    \vspace{-6.5mm}
    \caption{
    Qualitative comparison of dynamic object motion generation. CogVideoX~\cite{yang2024cogvideox} and Wan2.2-5B~\cite{wan2025wan} often produce distorted or unrealistic human motions with visual artifacts.
    HyperMotion~\cite{xu2025hypermotion} produces unnatural appearances because it relies on the first frame as input.
    In contrast, our method generates realistic videos with natural and highly dynamic motions.
    }
    \vspace{-1.5mm}
    \label{fig:qual}
\end{figure*}

\begin{table*}[!t]
\centering
\resizebox{\linewidth}{!}{
    \begin{tabular}{l|ccccc|ccccc}
    \toprule[1pt]
                    &           \multicolumn{5}{c|}{Pexels}                          &            \multicolumn{5}{c}{DynaVid-Human test}                           \\ \hline
                    & FVD ($\downarrow$)  & A-Qual ($\uparrow$) & I-Qual ($\uparrow$) & M-Smooth ($\uparrow$) & T-Flick ($\uparrow$) & FVD ($\downarrow$)  & A-Qual ($\uparrow$) & I-Qual ($\uparrow$) & M-Smooth ($\uparrow$) & T-Flick ($\uparrow$) \\ \hline
    CogVideoX-5B    & 1519.54              & 0.5646            & 0.6613            &  0.9844           & 0.9673                     & 2238.68           & 0.5071            & 0.5562         & 0.9779           & 0.9565           \\
    Wan2.2-5B          & 1172.02              & 0.5779            & 0.7235            & \textbf{0.9928}   & \textbf{0.9883}            & 1775.99           & \textbf{0.5389}   & 0.6974         & 0.9904           & 0.9791           \\
    HyperMotion        & 420.82*     & 0.5578             & 0.6972            & 0.9922          & 0.9850   & 391.22*  & 0.5092            & 0.6265         & \textbf{0.9939}  & \textbf{0.9914}  \\ 
    Ours     & \textbf{1126.38} & \textbf{0.5807}       & \textbf{0.7342}   & 0.9900            & 0.9748                     & \textbf{1351.94}  & 0.5312            & \textbf{0.7352} & 0.9931           & 0.9864           \\ 
    \bottomrule[1pt]
    \end{tabular}
}
\vspace{-3.0mm}
\caption{
Quantitative evaluation of object motion generation on the Pexels~\cite{pexels_dataset} and DynaVid-Human test datasets, representing common and highly dynamic scenes, respectively. Our method achieves comparable or superior results on both datasets compared to other baseline methods.
*: HyperMotion~\cite{xu2025hypermotion} uses a first frame as input, resulting in lower FVD scores due to the appearance similarity.
}
\vspace{-3.8mm}
\label{table:object_motion_evaluation}
\end{table*}

\section{Experiments}
\label{sec:experiment}

\vspace{-0.5mm}
\paragraph{Implementation details.}
We use Wan2.2-5B~\cite{wan2025wan} as the backbone for both the motion generator and the motion-guided video generator.
Following Wan2.2-5B, we synthesize 121 frames per sample at a resolution of 704$\times$1280 for both RGB frames and optical flow maps.
For the control branch architecture, we adopt VACE~\cite{jiang2025vace}.
Optical flow for real-world videos is extracted using WAFT~\cite{wang2025waft}, an off-the-shelf optical flow estimator.
Additional training details are provided in the supplementary document.

\vspace{-3mm}
\paragraph{Evaluation datasets and metrics.}
We evaluate our framework under two scenarios: object motion synthesis and camera-controlled video synthesis.
For each scenario, we prepare both real and synthetic datasets: the real datasets contain common scenes, while the synthetic datasets include highly dynamic motions.

For the object motion synthesis scenario, we use Pexels~\cite{pexels_dataset} for common scenes and the DynaVid-Human test set for highly dynamic scenes.
For the camera-controlled synthesis scenario, we use the RE10K test set~\cite{zhou2018stereo} for common scenes and the DynaVid-Camera test set for highly dynamic camera motion control.
We directly use the ground-truth camera parameters provided in the RE10K and DynaVid-Camera test sets.
From the real datasets (Pexels and RE10K), we randomly sample 100 video clips for evaluation.
For the synthetic datasets (DynaVid-Human and DynaVid-Camera), we additionally render 100 videos using our dataset generation pipeline (\cref{sec:syn_data_gen}).

We evaluate our models in terms of visual and motion quality, camera controllability, and motion fidelity.
For visual quality, we report Fréchet Video Distance (FVD)~\cite{unterthiner2018towards}, aesthetic quality (A-Qual), and imaging quality (I-Qual)~\cite{huang2024vbench}.
For motion quality, we measure motion smoothness (M-Smooth) and temporal flickering (T-Flick)~\cite{huang2024vbench}.
The four metrics, A-Qual, I-Qual, M-Smooth, and T-Flick, are computed using VBench~\cite{huang2024vbench}.
To assess camera controllability, we compute the mean rotation error (mRotErr) following CameraCtrl~\cite{he2025cameractrl}.
Finally, we evaluate motion error (M-Err) of the motion-guided video generator using the mean squared error between the input flow maps and those estimated from the generated video frames.

\begin{figure*}[!t]
    \centering
    \includegraphics[width=\linewidth]{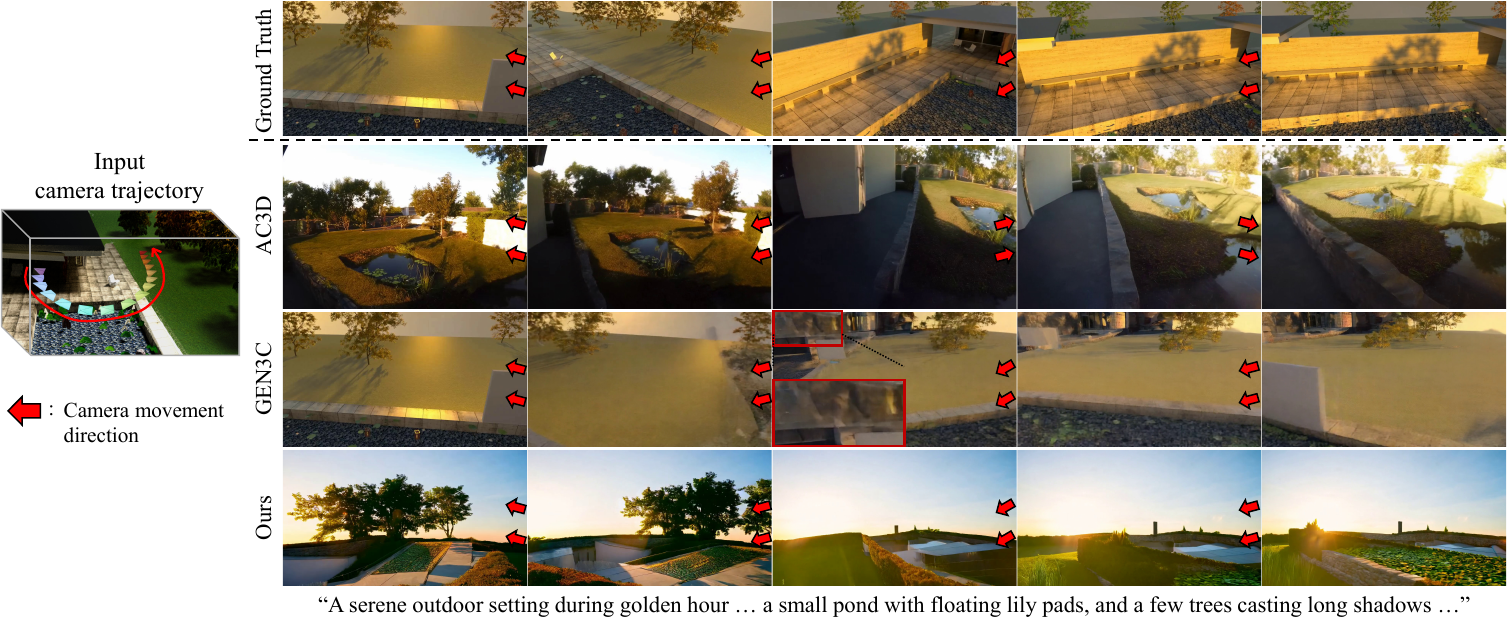}
    \vspace{-7mm}
    \caption{
    Qualitative comparison of camera-controlled video synthesis.
    Red arrows indicate the directions of camera motion.
    AC3D~\cite{bahmani2025ac3d} fails to follow the extreme 180$^{\circ}$ rotation, while GEN3C~\cite{ren2025gen3c} produces noticeable artifacts in regions unseen from the initial view (zoomed-in red box).
    In contrast, our method produces natural-looking videos that faithfully follow the input camera trajectory. For fair comparison, the same camera parameters are used for all methods.
    }
    \vspace{-2mm}
    \label{fig:qual_camera}
\end{figure*}

\begin{table*}[!t]
\centering
\resizebox{0.99\linewidth}{!}{
    \begin{tabular}{l|cccccc|cccccc}
    \toprule[1pt]
                    &           \multicolumn{6}{c|}{RE10K test}                 &                   \multicolumn{6}{c}{DynaVid-Camera test}                       \\ \hline
                    & mRotErr ($\downarrow$) & FVD ($\downarrow$)    &A-Qual ($\uparrow$) & I-Qual ($\uparrow$) & M-Smooth ($\uparrow$) & T-Flick ($\uparrow$)           & mRotErr ($\downarrow$)   & FVD  ($\downarrow$)  & A-Qual ($\uparrow$) & I-Qual ($\uparrow$) & M-Smooth ($\uparrow$) & T-Flick ($\uparrow$)           \\ \hline
    AC3D            & 0.1347              & 685.05            & 0.5184            & 0.6636            & \textbf{0.9918}   & 0.9651           & 1.1529           & 782.01          & 0.4483          & 0.5407          & 0.9727          & 0.9403           \\  
    GEN3C           & \textbf{0.0809}     & 566.99*           & 0.4579            & 0.6079            & 0.9899            & \textbf{0.9731}  & 1.1852           & 237.15*         & 0.3889          & 0.5659          & \textbf{0.9844} & \textbf{0.9611}  \\
    Ours            & 0.1136              & \textbf{664.21}   & \textbf{0.5425}   & \textbf{0.7224}   & 0.9888            & 0.9699           & \textbf{0.9289}  & \textbf{674.72} & \textbf{0.4501} & \textbf{0.6713} & 0.9760          & 0.9487           \\  
    \bottomrule[1pt]
    \end{tabular}
}
\vspace{-2mm}
\caption{
Quantitative evaluation of camera-controlled video synthesis on RE10K~\cite{zhou2018stereo} and DynaVid-Camera, representing common and dynamic scenes, respectively.
While achieving comparable rotation errors on common scenes, our method significantly outperforms other baseline methods on scenes with rapidly changing viewpoints.
*: GEN3C~\cite{ren2025gen3c} uses a first frame as input, resulting in lower FVD scores.
}
\vspace{-4mm}
\label{table:camera_control_evaluation}
\end{table*}

\subsection{Comparison on Dynamic Object Motion}
We compare our model with CogVideoX-5B~\cite{yang2024cogvideox}, Wan2.2-5B~\cite{wan2025wan}, and HyperMotion~\cite{xu2025hypermotion}.
CogVideoX and Wan2.2 are recent text-to-video generation models. 
HyperMotion is a Wan2.1-14B-based human-pose-controlled video generation method that takes a text prompt, a first frame, and a 2D human pose sequence as inputs, where the poses are extracted using an off-the-shelf pose estimator~\cite{yang2024x}.

\cref{fig:teaser,fig:qual} present qualitative comparisons on highly dynamic object motion generation.
While CogVideoX~\cite{yang2024cogvideox} and Wan2.2~\cite{wan2025wan} often fail to produce natural or dynamic movements due to the limited training data, our method successfully captures motion dynamics in highly dynamic video synthesis scenarios.
HyperMotion~\cite{xu2025hypermotion} tends to generate frames with an artificial appearance, partly because it relies on the first frame as a visual prior.

\cref{table:object_motion_evaluation} reports quantitative comparisons on datasets featuring common scenes (Pexels) and highly dynamic motions (DynaVid-Human).
Our method achieves comparable or superior visual and motion quality across both datasets compared to the baselines.
Although HyperMotion~\cite{xu2025hypermotion}, which leverages a larger backbone and additional input human pose sequences, performs comparably in dynamic scenarios, it exhibits degraded visual quality on common scenes, as it is primarily trained on human-centric video datasets.
Note that HyperMotion reports lower FVD scores as it directly uses the first frame as input, which increases appearance similarity with the evaluation datasets and consequently leads to lower FVD scores.

\subsection{Comparison on Extreme Camera Control}
We compare our model with AC3D~\cite{bahmani2025ac3d} and GEN3C~\cite{ren2025gen3c}.
Both AC3D and our method are text-to-video diffusion models that use Pl\"{u}cker embedding as control signals for camera-controlled video synthesis.
GEN3C, in contrast, is an image-to-video diffusion model that estimates depth from the input image to reconstruct 3D caches (e.g., point clouds) for 3D-aware camera control.

\cref{fig:teaser,fig:qual_camera} present qualitative comparisons with the baseline methods.
AC3D fails to follow extreme camera trajectories involving rapid viewpoint changes (e.g., 180$^{\circ}$ rotations), while GEN3C produces artifacts in regions unseen from the input image.
In contrast, our method generates realistic frames that accurately follow challenging camera paths, benefiting from training with synthetic motion data.

\cref{table:camera_control_evaluation} reports quantitative comparisons with the baseline models.
While our model shows comparable camera controllability on RE10K~\cite{zhou2018stereo}, which includes moderate viewpoint changes, it significantly outperforms both baselines on DynaVid-Camera, which contains rapidly changing camera trajectories.
Moreover, our method demonstrates higher visual quality across the datasets.
Additional qualitative examples are provided in the supplementary document.

\begin{table*}[!t]
\centering
\resizebox{\linewidth}{!}{
    \begin{tabular}{l|ccccc|ccccc}
    \toprule[1pt]
                    &           \multicolumn{5}{c|}{Pexels}                          &            \multicolumn{5}{c}{DynaVid-Human test}                           \\ \hline
                    & FVD ($\downarrow$)  & A-Qual ($\uparrow$) & I-Qual ($\uparrow$) & M-Smooth ($\uparrow$) & T-Flick ($\uparrow$) & FVD ($\downarrow$)  & A-Qual ($\uparrow$) & I-Qual ($\uparrow$) & M-Smooth ($\uparrow$) & T-Flick ($\uparrow$) \\ \hline
    Ours            & 1126.38    & 0.5807   & 0.7342    & 0.9900    & 0.9748       
                    & 1351.94    & 0.5312   & 0.7352    & 0.9931    & 0.9864  \\ \hline
    \hspace{1mm}w/o synthetic motion data (a)
                    & \cellcolor{greenLight}(-49.85)
                    & \cellcolor{redStrong}(-0.0199)
                    & \cellcolor{redStrong}(-0.0156)
                    & (-0.0002)
                    & \cellcolor{redLight}(-0.0019)
                    & \cellcolor{redStrong}(+527.04)
                    & \cellcolor{greenStrong}(+0.0106)
                    & \cellcolor{redStrong}(-0.0403)
                    & \cellcolor{redLight}(-0.0062)
                    & \cellcolor{redStrong}(-0.0249) \\
    \hspace{1mm}w/o batch mixture (b)
                    & \cellcolor{redStrong}(+759.36)
                    & \cellcolor{redLight}(-0.0073)
                    & \cellcolor{redLight}(-0.0070)
                    & \cellcolor{greenLight}(+0.0030)
                    & \cellcolor{greenStrong}(+0.0144)
                    & \cellcolor{greenLight}(-122.24)
                    & \cellcolor{redLight}(-0.0050)
                    & (-0.0006)
                    & \cellcolor{redLight}(-0.0022)
                    & \cellcolor{redLight}(-0.0013) \\
    \hspace{1mm}w/o dataset filtering$^{\dagger}$ (c)
                    & \cellcolor{greenLight}(-33.79)  
                    & \cellcolor{greenLight}(+0.0011) 
                    & \cellcolor{greenLight}(+0.0035) 
                    & (+0.0007) 
                    & (+0.0009) 
                    & \cellcolor{redLight}(+32.44)  
                    & \cellcolor{greenLight}(+0.0055) 
                    & \cellcolor{redStrong}(-0.0140) 
                    & (+0.0008) 
                    & \cellcolor{greenLight}(+0.0012) \\
    \hspace{1mm}w/~~ synthetic video data (d)
                    & \cellcolor{redStrong}(+104.43)
                    & \cellcolor{greenStrong}(+0.0163)
                    & \cellcolor{greenStrong}(+0.0124)
                    & \cellcolor{redLight}(-0.0012)
                    & (+0.0005)
                    & (-653.98)*
                    & \cellcolor{greenStrong}(+0.0112)
                    & \cellcolor{redStrong}(-0.0487)
                    & (-0.0000)
                    & \cellcolor{greenLight}(+0.0049) \\
    \bottomrule[1pt]
    \end{tabular}
}
\vspace{-3.5mm}
\caption{
Ablation study of our main components, evaluating video and motion synthesis quality on the Pexels and DynaVid-Human test datasets.
$\dagger$: Our model without data filtering achieves comparable quantitative results but produces noticeably degraded visual quality in highly dynamic motion synthesis, as shown in~\cref{fig:ablation}.
*: Our model trained with rendered synthetic videos reports lower FVD scores, because it produces artificial appearances resembling synthetic data, as shown in~\cref{fig:ablation}.
}
\vspace{-4.5mm}
\label{table:ablation}
\end{table*}

\subsection{Analysis}
\label{sec:ablation}

\vspace{-0.5mm}
\paragraph{Importance of synthetic motion data.}
We first investigate the importance of synthetic motion data in training, which provides highly dynamic motion patterns to the model~(\cref{sec:training}).
As shown in~\cref{table:ablation}, removing synthetic motion data leads to a significant degradation in both visual and motion quality on DynaVid-Human, resulting in much higher FVD scores for highly dynamic video generation. In contrast, only a minor change in FVD is observed on Pexels.

\vspace{-3.5mm}
\paragraph{Importance of batch mixture.}
During fine-tuning of the motion generator, we employ mixed training batches containing both real and synthetic flows~(\cref{sec:training}).
To analyze the effect of this strategy, we train a variant that uses only synthetic motion data during fine-tuning, excluding real data from the batch mixture.
As reported in~\cref{table:ablation}, this variant exhibits substantial performance degradation with FVD scores on the Pexels dataset.
This indicates that real flow data is necessary to preserve general motion priors, while training with synthetic flow data alone leads to overfitting to synthetic motion patterns.

\begin{figure}[!t]
    \centering
    \includegraphics[width=\linewidth]{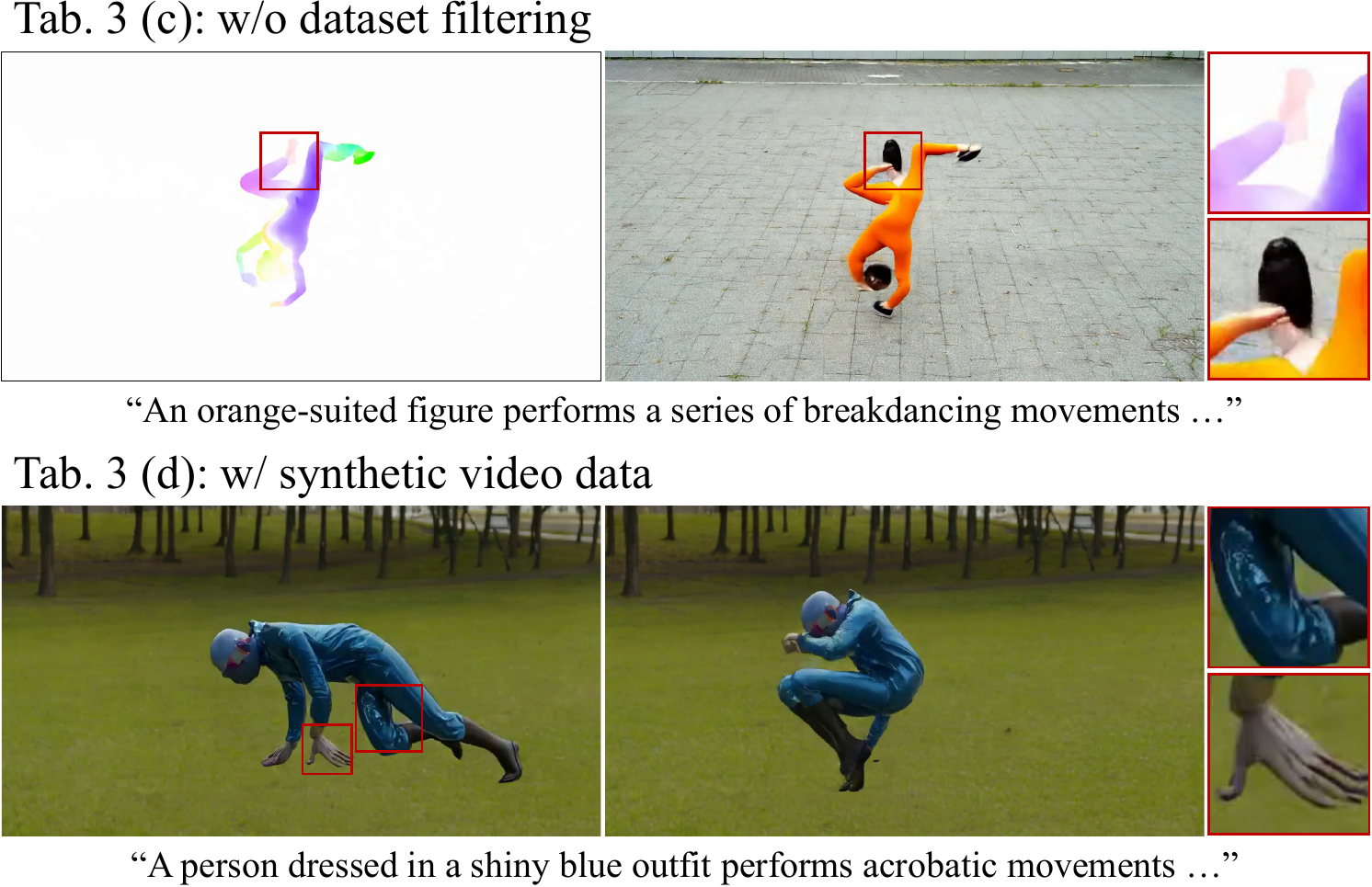}
    \vspace{-6.0mm}
    \caption{
    Visual examples of the ablation study. Our model without dataset filtering produces video frames that are inconsistent with the input motion (top).
    Our model trained using the synthetic video dataset reproduces the artificial look of rendered videos (bottom).
    }
    \vspace{-4.0mm}
    \label{fig:ablation}
\end{figure}

\vspace{-4.0mm}
\paragraph{Importance of dataset filtering.} 
We apply a simple dataset filtering technique based on flow cycle consistency to reduce optical flow estimation errors in the real dataset~(\cref{sec:training}).
To evaluate its effect, we train a motion-guided video generator without filtering.
Although this variant shows similar quantitative results~(\cref{table:ablation}), it produces unnatural videos in highly dynamic scenarios.
As shown in~\cref{fig:ablation}, this model generates misaligned body parts (e.g., incorrect head positions highlighted in red), revealing inconsistency between video frames and input motions.
This is also validated with the motion error (M-Err), where the variant reports 4.734 compared to 4.287 for our final model.

\vspace{-4.0mm}
\paragraph{Effect of using rendered videos.}
We further examine the impact of using rendered synthetic videos instead of synthetic motion data, which 
represents the most straightforward way to leverage synthetic datasets.
To this end, we train an additional model that uses rendered videos together with batch mixture training on real videos.
As shown in~\cref{table:ablation}, this model exhibits degraded performance on Pexels, reporting higher FVD scores.
Moreover, despite the use of batch mixture, it produces videos with an artificial appearance as shown in~\cref{fig:ablation}, which leads to lower FVD scores on DynaVid-Human due to the synthetic visual characteristics of that dataset.
These results demonstrate the limitations of directly using rendered videos for training.

\vspace{-3.5mm}
\paragraph{Robustness to motion generation error.}
Because the motion-guided video generator directly takes the motion synthesized by the motion generator as input, it can be affected by motion generation errors.
Thus, we validate whether the motion-guided video generator can synthesize high-quality video frames while faithfully following motion when the input flow is noisy.
We prepare clean optical flow and noisy variants by adding Gaussian noise to achieve target signal-to-noise ratios of 25, 20, 15, 10, and 5 dB.
Then, the motion-guided video generator synthesizes videos conditioned on each flow variant.
\cref{table:analysis_robustness} reports the motion error and aesthetic quality.
Our framework remains robust at 20dB and above, with gradual degradation as the noise level increases.
Further experimental details are provided in the supplementary document.

\begin{figure}[!t]
    \centering
    \includegraphics[width=\linewidth]{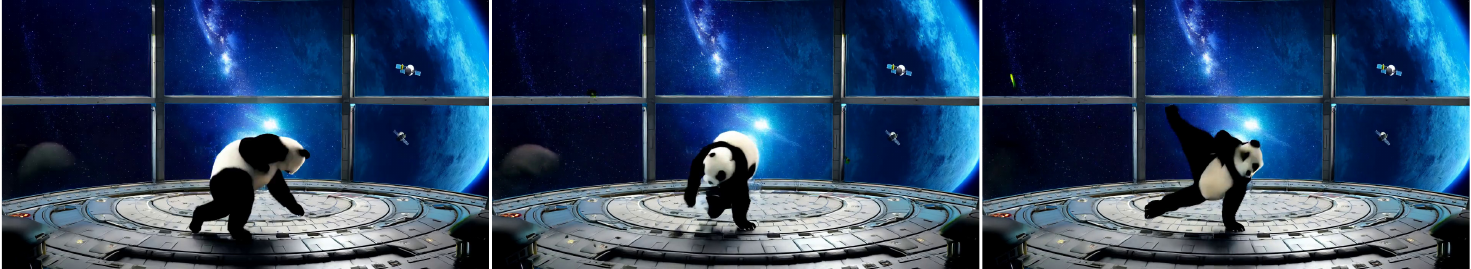}
    \vspace{-7mm}
    \caption{
    Capability to synthesize non-human dynamic motions.
    }
    \vspace{-3.mm}
    \label{fig:analysis_generalization}
\end{figure}

\vspace{-3.5mm}
\paragraph{Behavior beyond the training domain.}
Although the DynaVid-Human dataset contains only single-person scenes, our method can synthesize motions that extend beyond this training domain.
As shown in~\cref{fig:analysis_generalization}, the model generates a dynamically moving panda whose body structure differs from that of humans, despite being trained solely on human-centric data.
This indicates that our approach can produce plausible motions for certain unseen object types, even though it is not explicitly trained for such cases.

\begin{table}[!t]
\centering
\resizebox{\columnwidth}{!}{
    \begin{tabular}{l|cccccc}
    \toprule[1pt]
                                & Clean     & 25dB      & 20dB      & 15dB     & 10dB     & 5dB     \\ \hline
    M-Err ($\downarrow$)   & 4.287     & 4.371     & 4.374     & 4.323    & 4.391    & 4.434   \\
    A-Qual ($\uparrow$)        & 0.5312    & 0.5233    & 0.5239    & 0.5148   & 0.5071   & 0.4632  \\
    \bottomrule[1pt]
    \end{tabular}
}
\vspace{-3.0mm}
\caption{
Error robustness of the motion-guided video generator.
}
\vspace{-3.0mm}
\label{table:analysis_robustness}
\end{table}

\begin{figure}[!t]
    \centering
    \includegraphics[width=\linewidth]{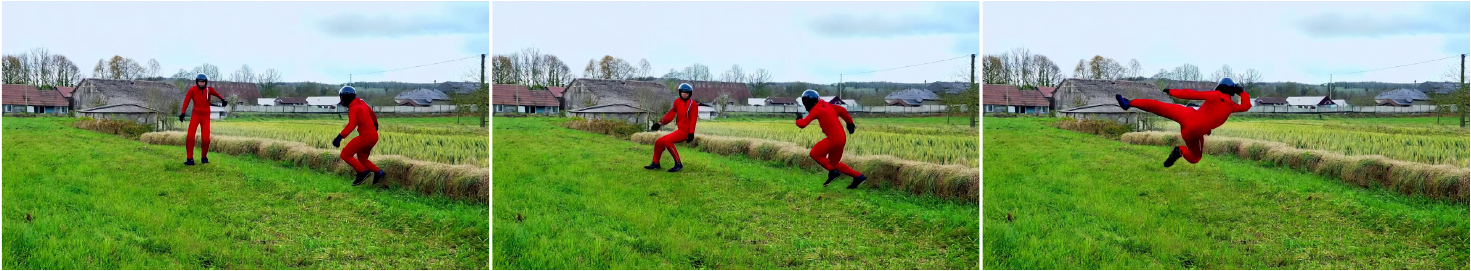}
    \vspace{-7mm}
    \caption{
    Limitations. Our method is less effective in generating videos with multiple people performing dynamic motions.
    }
    \vspace{-3mm}
    \label{fig:limitation}
\end{figure}

\vspace{-3.5mm}

\paragraph{Limitations.}
Our method is not free from limitations.
Since our synthetic dataset primarily consists of single-person scenes, the model often struggles to generate videos involving multiple people with highly dynamic motions~(\cref{fig:limitation}).
This issue could be mitigated by increasing the diversity and scale of the synthetic dataset, which we leave for future work.
Additional discussions of these limitations are provided in the supplementary document.

\section{Conclusion}
\label{sec:conclusion}

We presented \MethodName{}, a video synthesis framework for generating realistic videos with highly dynamic motions and controllable camera movements.
Our key idea is to leverage \emph{synthetic motion data} to provide dynamic and precise motion supervision while avoiding appearance-domain gaps.
Through a two-stage design that separately models motion generation and motion-guided video synthesis, our method effectively leverages synthetic and real-world data to achieve both visual realism and dynamic controllability.
Extensive experiments demonstrate that DynaVid outperforms existing video diffusion models in dynamic motion generation and camera-controlled synthesis, while maintaining robustness to motion generation errors and moderate out-of-domain scenarios.

\paragraph{Acknowledgment.}
This work was supported by the Institute of Information \& Communications Technology Planning \& Evaluation (IITP) grant funded by the Korea government (MSIT) under the following programs: RS-2024-00395401 (Development of VFX creation and combination using generative AI), RS-2019-II191906 (Artificial Intelligence Graduate School Program at POSTECH), RS-2024-00457882 (AI Research Hub Project), and RS-2024-0045788. This work was also supported by the National Research Foundation of Korea (NRF) grant funded by the Korea government (MSIT) (RS-2023-00211658 and RS-2024-00438532).
This work was also supported by Microsoft Research Asia.
{
    \small
    \bibliographystyle{ieeenat_fullname}
    \bibliography{main}
}


\end{document}